\pdfoutput=1
\documentclass[11pt]{article}

\usepackage[]{acl}

\usepackage{times}
\usepackage{latexsym}
\usepackage{url}
\usepackage{color}
\usepackage{amsmath}
\usepackage{amsfonts}

\newcommand{\cready}[1]{\textcolor{pink}{{}}}

\usepackage[T1]{fontenc}
\usepackage[utf8]{inputenc}

\usepackage{microtype}

\usepackage{graphicx}
\graphicspath{ {./images/} }

\title{PreQuEL: Quality Estimation of Machine Translation Outputs \\ in Advance}

\author{Shachar Don-Yehiya \qquad Leshem Choshen  \qquad Omri Abend \\
  School of Computer Science and Engineering, The Hebrew University of Jerusalem \\
  \texttt{\{first.last\}@mail.huji.ac.il} \\}
  
\begin{document}
\maketitle

\begin{abstract}
%The use of machine translation to substitute human translators is getting increasingly common. However, MT systems can fail unexpectedly. This raises the question of predicting whether a given source sentence (or corpus) should be translated or not. 
We present the task of {\it PreQuEL}, Pre-(Quality-Estimation) Learning. A PreQuEL system predicts how well a given sentence will be translated, without recourse to the actual translation,
%While the task of Quality Estimation focuses on determining whether a sentence is a correct translation of a source sentence, i
%We introduce the task of ``Pre-Quality Estimation" task, the task of predicting the quality of the output of Neural Machine Translation systems without access to the actual output.
%This task determines whether it is likely to achieve a high-quality automatic translation, before investing the necessary resources to actually translate.
thus eschewing unnecessary resource allocation when translation quality is bound to be low. 
PreQuEL can be defined relative to a given MT system (e.g., some industry service) or generally relative to the state-of-the-art.
% Are there any specific properties, maybe linguistic features, that make a sentence more difficult for MT? By introducing the PreQuEL task, we aim to place the focus on the input text, and use the MT system to learn which source texts are easy to translate and which are difficult
%\sdy{Additionally, this task has a theoretical value. by placing the focus on the input text, providing insight into the performance boundaries of state-of-the-art MT methods.}
From a theoretical perspective, PreQuEL places the focus on the source text, tracing properties, possibly linguistic features, that make a sentence harder to machine translate.

We develop a baseline model for the task and analyze its performance. We also develop a data augmentation method (from parallel corpora), that improves results substantially. We show that this augmentation method can improve the performance of the Quality-Estimation task as well.\footnote{Code and data are available in: \url{https://github.com/shachardon/PreQuEL}} %We further reinterpret past findings about Quality-Estimation evaluation, in light of this study.
%To boost our model performance, we introduce a simple method to create pseudo Quality Estimation and Pre-Quality Estimation data from parallel corpora and automatic evaluation metrics.
%%%%
We investigate the properties of the input text that our model is sensitive to, by testing it on challenge sets and different languages. We conclude that it is aware of syntactic and semantic distinctions, and correlates and even over-emphasizes the importance of standard NLP features. %Our model manages to distinguish between translation from one source language into different target languages, although it never explicitly sees the target languages.
\end{abstract}

%%%%%%%%%%%%%%%%%%%%%%%%%%%%%%%%%%%%%%%%%%%%%
\section{Introduction}

%\oa{We examine the advantages of advanced architectures, to test the possible effect of syntax knowledge or morphological features.}
%\oa{It is also important from a theoretical standpoint: mapping the boundaries of MT systems.}

Can we tell if a sentence is difficult to automatically translate, without actually translating it? We argue that this question has important practical and theoretical value. 

\begin{figure}[t!]
\centering
\includegraphics[width=7.6cm]{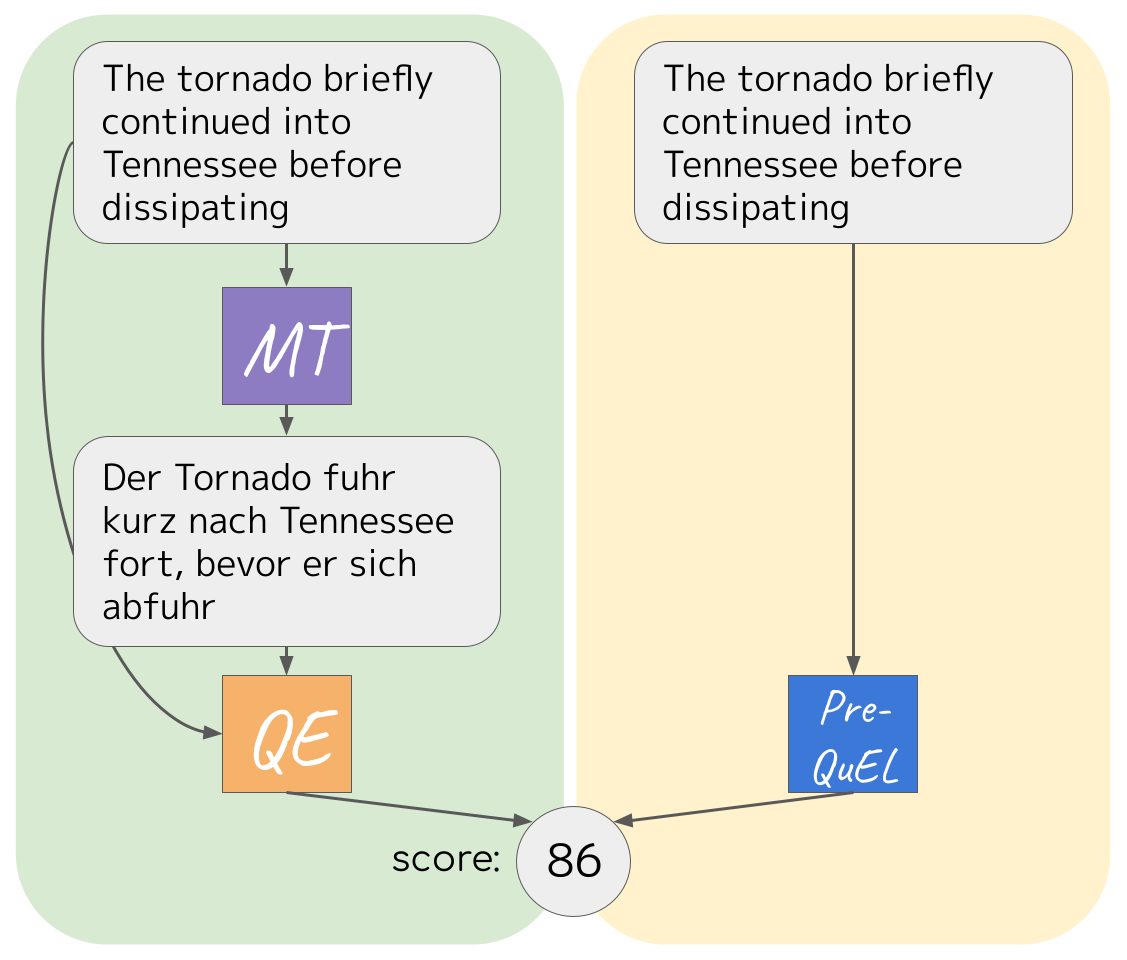}
\caption{Quality Estimation (QE) vs. PreQuEL. In the PreQuEL task, the score is generated from the source sentence directly, before investing the necessary resources to translate.}
\label{figure:prequel-vs-qe}
\end{figure}

%From a practical standpoint, it can save the trouble and cost of generating Machine Translation (MT) outputs, and focus human translation efforts where they are needed.
From a practical standpoint, a PreQuEL system can save trouble and cost.
%\sdy{Maybe something like: From a practical standpoint, a PreQuEL system can generate a quality score before investing the resources to actually translate, saving the trouble and cost of generating going-to-be low-quality Machine Translation (MT) outputs.}
For individuals or companies who do not maintain their in-house Machine Translation (MT) system, translating a big amount of data might turn out to be expensive. There are several automatic translation services offered, but there is no prior indication that any of them would do a good enough job. Therefore, after purchasing translations, one would probably want to estimate the quality of the translations, with human annotators or an automatic Quality Estimation (QE) system \citep{blatz-etal-2004-confidence, specia-etal-2009-estimating}.
After spending these resources, there is still a chance that the translation quality will be judged unsatisfying, and there will be no choice but to hire human translators.  
The decision between the automatic route and the much more expensive alternative of hiring human translators can be informed by a PreQuEL model, before investing the resources to actually translate. See figure \ref{figure:prequel-vs-qe}.

%\footnote{\oa{I think we need to say a bit more; why is this an important step relative to QE?} \sdy{In practical or theoretical terms? if theoretical, maybe we should talk about the model as an explanatory tool for MT.}\oa{Both: theoretically, as you say, it's about the boundaries of MT. Practically, we should say why this is so demanding to actually run these engines. To some it might look like: well, it's just inference, and the code is there.}} 

Theoretically, PreQuEL allows insight into the performance boundaries of state-of-the-art MT methods.
It is common practice to search in retrospect what kinds of text phenomena are difficult for existing MT systems to handle \citep{toral-sanchez-cartagena-2017-multifaceted}\cready{Muler?}. Such analysis helps in finding flaws in the system, and getting some notion of what is needed to overcome them. Challenge sets are also used for that purpose -- testing the system, analyzing its performance on carefully chosen cases \citep{isabelle-etal-2017-challenge, barrault-etal-2020-findings,choshen-abend-2019-automatically}. However, no attention was heeded to the source sentence itself and how it affects the translation quality. Are there any specific properties, maybe linguistic features, that make a sentence more difficult for MT? By introducing the PreQuEL task, we place the focus on the input text, and learn which source texts are easy to translate and which are difficult.

% There are cases when there is a need to determine the chances for achieving a good quality automatic translation before investing the necessary resources to actually translate. Imagine your company has a huge English corpus, which needs to be translated to Chinese. They are several automatic translation services that you can pay to, but there is no prior indication that any of them would make a good enough job. The decision between the automatic route and the much more expensive alternative of hiring human translators can be informed by a PreQuEL model. %In a case like this, a Pre-QE model can help to better estimate the chances before taking the decision.  

We consider two variants of this approach. One, where the system is given 
%(where a sufficient number of sentences for its training is available), 
and another where the prediction is done relative to the state-of-the-art in the field. The latter of course bears a tacit assumption that there are underlying properties that make a sentence easier or harder to translate for a state-of-the-art system. Supporting empirical evidence for this assumption is presented in \S\ref{Generalizing to Other MT Systems}.

%\oa{add here a paragraph about the methods and about the results. Telegraphically. the details will go into further sections.}

We develop a baseline model for the task, based on a state-of-the-art QE system, and report its results in \S\ref{results}. We further develop an automatic data augmentation method from parallel corpora and use it in intertraining \citep{Phang2018SentenceEO} and multi-task settings, outperforming our initial results (\S\ref{Automatic Evaluation Metrics Data}). In \S\ref{QE}, we show that this augmentation method can improve the performance of the QE task as well. Our results, surprisingly, do not fall far behind the results of state-of-the-art QE systems, despite not being exposed to the actual translation.

%To further investigate the function that our model learns, we test it on a word-ordering challenge set. We conclude that our model predicts higher scores for sentences common word order sentences, and gives similar scores for sentences that differ in their syntax but are identical in their meaning. Following these results, we suggest a method for using our model as an explanatory tool for MT. We compare the predictions of a model that was trained for English-German with a model that was trained for English-Chinese. The comparison shows the model manages to learn different functions for different target languages, although the input to the model does not specify the target explicitly.

To the best of our knowledge, this work is the first to address PreQuEL. \citet{sun-etal-2020-estimating} did point out that it is possible to perform QE 
%to a large extent 
using only the input sentence, but viewed this finding as an artifact of their dataset. We revisit their claims and argue that the PreQuEL model can potentially simulate the MT system, and there is thus no theoretical reason to consider the dataset as ``cheatable''. We discuss their claims extensively in \S\ref{QE}.

Analyzing the proposed model, we examine in \S\ref{Generalizing to Other MT Systems} the question of whether the predictions made by the model are specific to the performance of one particular system (whose outputs were used for supervision), or whether they generalize to other recent neural systems. Our results indicate that results do generalize to other systems, finding a drop of correlation of only $4.2$ points.

We further examine the contribution of the syntactic structure of the output in predicting its difficulty. To do so we experiment with a model that was fine-tuned with syntactic parsing, but find that its improvement to the performance is relatively low (\S\ref{Advanced Models}). However, we do find a correlation between the baseline model's predictions and some syntactic features, suggesting that the model is aware of syntax (\S\ref{word ordering}). We further explore what features of the input sentence our model is sensitive to, and report results in \S\ref{standard features}.

% using the Q) data to train and test our model. The goal of the QE task is to estimate the quality of neural machine translation (MT) output without relying on a reference translation. In the Pre-QE task, we take another step - we estimate the quality of the output without relying on the output itself.

%%%%%%%%%%%%%%%%%%%%%%%%%%%%%%%%%%%%%%%%%%%%%
\section{Task Definition}

PreQuEL is the task of predicting the likelihood of an MT system to correctly translate a given sentence to a given target language.
The task has two variants. 

The system-specific variant is given a source language $S$, a target language $T$ and an MT system between them $M$. The goal is to learn a function $g: S \rightarrow \mathbb{R}$, such that for every sentence $s\in S$ the score $g(s)$ represents the expected quality of the translated sentence $M(s)=t\in T$.
As standard in QE evaluation, the PreQuEL predictions are evaluated against the gold labels using Pearson's $r$ correlation coefficient \citep{specia-EtAl:2020:WMT2}.

The definition for the second variant, namely of general state-of-the-art MT, is equivalent but with no specific $M$. Instead, $g$ satisfies the above for all $M$ such that $M$ is a state-of-the-art MT. Unlike the first, the correlation achievable in this variant is inherently bound, due to the non-uniform behavior exhibited by different MT systems. Nevertheless, to the extent that systems do share performance patterns, mapping them is of theoretical and applicative value. 

We focus on the first variant at first and examine the second in \S\ref{Generalizing to Other MT Systems}.

%%%%%%%%%%%%%%%%%%%%%%%%%%%%%%%%%%%%%%%%%%%%%%
\section{Data}

\subsection{Quality Estimation Data}

The WMT shared task on QE includes estimation at three granularity levels: word, sentence, and document. The WMT 2020 \citep{specia-EtAl:2020:WMT2} introduced a variant of the sentence-level task. The sentences were annotated with direct assessment (DA) scores \citep{graham-etal-2013-continuous},
%\lc{was this due to the findings we mention at the end? If so add something like: we revisit the reasons for this move, reaching different conclusions.} \sdy{obviously yes, but they do not say it explicitly in the shared task paper. when they do cite this paper, it's for something else. What do you think?} 
instead of labels based on post-editing \citep{snover-etal-2006-study}. The dataset for the task is composed of data extracted from Wikipedia for six language pairs, of which we use the high-resource languages English-German (en-de) and English-Chinese (en-zh) and medium-resource pair Estonian-English (et-en). Each language pair has 7K,1K,1K sentence pairs in the training, development and test sets respectively. Translations were produced with a state-of-the-art MT model built using the fairseq toolkit \citep{ott-etal-2019-fairseq}. %Each translation was rated with a score between 0 and 100 following the FLORES guidelines \citep{guzman-etal-2019-flores} by at least three professional translators \citep{fomicheva2021mlqepe}. The DA scores were standardized using the z-score. For our purposes, we take only the source sentences and the DA scores, ignoring the translations. To facilitate training, we transform the z-scores to a 0-1 scale with a min-max normalization. 
Each translation was rated following the FLORES guidelines \citep{guzman-etal-2019-flores}. For our purposes, we take only the source sentences and the DA scores, ignoring the translations. To facilitate training, we transform the scores to a 0-1 scale with a min-max normalization.

\subsection{Augmented Data}
\label{Automatic Evaluation Metrics Data}

We propose an automatic method to acquire more translation quality scores, dispensing with the costly human annotations.

Automatic evaluation metrics \citep{mathur2020results} give additional information on translation quality. Unlike QE, such metrics are exposed to human translation as well. While relying on QE for augmentation would possibly be beneficial, we opt for extracting the latent information found in automatic metrics instead.
% Relying on QE for augmentation would be bootstrapping, while possibly still beneficial, we opt for extracting the latent information found in automatic metrics instead.

Given a parallel corpus, we re-translate it with the MT system $M$. We take the source sentences to be the PreQuEL inputs, and the metric scores to be the labels.

Where automatic metrics are often far from agreeing with human judgments \citep{choshen-abend-2018-inherent, kocmi-etal-2021-ship}, they still extract some potentially beneficial aspects of similarity to the translation. As the PreQuEL has no access to those references (or translations), training to predict the automatically estimated quality of the augmented data encourages the PreQuEL system to extract those features.

We examine the difference between predicting DA and metric scores directly in appendix \S\ref{DA vs. COMET}. Nonetheless, we show it is similar enough to be beneficial as a data augmentation method in \S\ref{results}.

%%%%%%%%%%%%%%%%%%%%%%%%%%%%%%%%%%%%%%%%%%%%%%%%%%%%%%%%%%
\section{Models}

\subsection{Baseline Model}\label{Baseline}

We propose a baseline system model for PreQuEL. Our implementation builds on TransQuest  \citep{ranasinghe-orasan-mitkov:2020:WMT}, the winner system of the WMT 2020 QE sentence-level DA task \citep{specia-EtAl:2020:WMT2}. The model is builds on RoBERTa-large \citep{liu2019roberta} to derive the representations of the input sentence. For pooling it uses the output of the [CLS] token.
%with a fully connected layer on top. 
After pooling we place a 1024$\times$1024 and 1024$\times$1 fully-connected layers. The latter produces the final output.%\oa{add an appendix with all the nitty gritty hyperparams and reference it from here}
 
\subsection{Advanced Models}\label{Advanced Models}

We examine the advantages of more advanced architectures.
%\sdy{
These architectures combine syntactic or semantics components, that may improve PreQuEL results. Additionally to that, our other motivation is to get a better understating of PreQuEL required features. In the case of a significant improvement over the baseline models, we would conclude that syntactic and semantic knowledge are crucial for PreQuEL.
%}

%More than improving PreQuEL results, such models combine syntactic or semantics components that when compared to the baseline models can give a better explanation of the function that our model learns.\oa{is this the motivation? is this what the results show? the last sent s.b. rephrased}
 
\paragraph{\textsc{Combined}.}

To examine the need for external syntactic knowledge, we combine a UD parser with our architecture. We use RobertNLP \citep{grunewald-friedrich-2020-robertnlp}, which is also RoBERTa based. Using the parser as an intertraining step resulted with lower performance. Instead, we concatenate the last hidden layer of a fresh pre-trained RoBERTa model and that of the parser. The concatenated hidden layers are used as the input to the classifier (same as the one in the original architecture, with adjustments to the dimensions).

\paragraph{\textsc{Multitask}.}
\label{Multi-task Mode}
Under certain circumstances, related tasks may help each other using Multi-task Learning (MTL) \citep{aghajanyan-etal-2021-muppet}. As discussed in \S\ref{Automatic Evaluation Metrics Data}, the task of predicting other automatic metrics might help predict DA too. We extend the model with additional classifier heads, each predicting a different automatic metric. 
\section{Experimental Setup}
\label{exp setup}

We train and test our PreQuEL models each time on one language pair.
The first pair we examine is the high-resource en-de. As discussed in \citet{fomicheva2021mlqepe}, the translation quality for translations in this pair has little variability, with a mean score of $73.25$ and std $8.13$. The vast majority of translations were assigned high DA scores, which makes differentiating between them challenging.

To balance this, we select as our second pair the medium-resource et-en. Non-high-resource pairs give QE models an advantage -- outputs are occasionally 'hallucinated', i.e., they do not have anything to do with the original sentences. Detecting such cases should be simple for QE systems, which explains the high QE scores on those pairs. We would therefore like to examine whether this effect persists in the PreQuEL settings, where the outputs are not available to the model.

We use the pipeline described in \S\ref{Automatic Evaluation Metrics Data} to create scores for en-de and en-zh (when available) WMT-News \citep{barrault-etal-2020-findings}, bible-uedin \citep{Christodoulopoulos2015AMP}, Tatoeba \citep{tiedemann-2020-tatoeba} and GlobalVoices \citep{TIEDEMANN12.463}, all taken from OPUS \citep{TIEDEMANN12.463}.
We remove duplicate sentences and randomly split each of the datasets to train/dev/test, 80\%/10\%/10\%. For the MT system we use the OPUS-MT released model \citep{TiedemannThottingal:EAMT2020}, based on Marian-MT \citep{mariannmt}.

The selected datasets are diverse in their domain. WMT-News (28,887 sentences) is a parallel corpus of news provided by WMT for testing MT performance. bible-uedin (48,705 sentences) is a multilingual parallel corpus created from translations of the Bible. Tatoeba (197,381 sentences) is a crowd-sourced collection of user-provided translations. GlobalVoices (55,822 sentences) is a news parallel corpus from the website \href{https://globalvoices.org/}{Global Voices}. We examine the out-of-domain effect in App.~\S\ref{out-of-domain}. We also ensure that for each language pair at least two datasets were not found in Marian training set, making PreQuEL predictions more interesting.

We note that despite we translate from English to German/Chinese, the translation direction is not always from English. This might lead to artifacts in the metric evaluation \cite{graham-etal-2020-statistical}. However, we speculate this is more of a concern for evaluation. Most of our experiments rely on the data as an augmentation method, which either helps the main task or not.

For the automatic metric we use COMET \citep{rei-etal-2020-comet}. Where more than one metric is required (for training \textsc{Multitask}), we use also ChrF++ \citep{popovic-2015-chrf} and BERTScore \citep{bert-score}. These three metrics cover different kinds: ChrF++ is string-based, BERTScore is an unsupervised embedding-based model, and COMET is a supervised model trained end to end. 

% \begin{table}[t!] \footnotesize
% \centering
%     \begin{tabular}{ c|c|c|c } 
%     \hline
%     {NewsTests} & {bible-uedin} & {GlobalVoices} & {Tatoeba} \\
%     \hline
%     {28,887} & {48,705} & {55,822} & {197,381} \\
%     \hline
%     \end{tabular}
% \caption{Number of sentences in the train set for each dataset.\oa{I don't think there is justification for a single row table. You can put these numbers inthe text.}}
% \label{table:1}
% \end{table}

%, see appendix \ref{metrics implementation}.
%\footnote{\url{https://github.com/m-popovic/chrF, https://unbabel.github.io/COMET/html/index.html, https://github.com/Tiiiger/bert_score}}.

We train instances of \textsc{Simple} for en-de and et-en.
For en-de and en-zh we train also a \textsc{Simple} Aug version with COMET as intertraining. 
%We train \textsc{Simple} with COMET as intertraining \citep{Phang2018SentenceEO}, '\textsc{Simple} Aug', for en-de and en-zh.
For the en-de en-zh comparison (\S\ref{Different Target Languages}), we use datasets that are available in both languages: NewsTests and bible-uedin, and test them on the subset of source sentences that are shared between en-de and en-zh development DA.
%To provide more translations from languages other than English, we train a version of \textsc{Simple} on de-en COMET, '\textsc{Simple} COMET'. 
For en-de we also train the more advanced architectures, \textsc{Combined} and \textsc{Multitask}. We train two versions of \textsc{Combined}, \textsc{Combined+} and \textsc{Combined- }, the first concatenated to an actual UD parser and the latter to another pre-trained RoBERTa to control for the size. We use the augmentation intertraining for both versions. Where we compare two or more versions, we train 3 seeds for each. 

We take the hyperparameters from the TransQuest implementation, Adam optimizer with a learning rate $1e-5$, and a linear learning rate warm-up over 10\% of the training data. We adjust the batch-size from 8 to 4 to fit our GPUs. See App.~\ref{Computing appendix}.

%We use a batch-size of 4, Adam optimizer with a learning rate $1e-5$, and a linear learning rate warm-up over 10\% of the training data. We take the hyperparameters from the Transquest implementation, \sdy{except for the batch-size which we adjust due to the limited resources we have.} adjusting them to the limited resources we have\oa{what does that mean in practice?}. 
%During the training process, the parameters of the RoBERTa-large model \footnote{\url{https://huggingface.co/roberta-large}}, as well as the parameters of the subsequent layers, are updated. 
To allow evaluation during training, we sampled 10\% of the training data and kept it for evaluation. This is in addition to the development data we use for evaluation at the end of the training. We carry out evaluation every 300 training steps for a small training set (smaller than 1K steps), and every 3K steps otherwise. We perform early stopping with patience of 10 evaluation rounds. The model is trained for a maximum of 3 epochs, and no less than one. If we early-stop during the first epoch \citep{Dodge2020FineTuningPL}, or reach a low correlation ($<0.1$) on the last evaluation round, we reset the seed controlling initialization, batches, split of the training/evaluation data, etc. In order to improve our results, we train 3 random seeds and infer by an average ensemble.
We carry out evaluation on the DA test set.

%In places where we compare ourself to a QE system, we use TransQuest.%\oa{this sentence is rather confusing. I would include all the material relevant to the QA evaluation in that specific section}
%\footnote{We use the code from their github. We did not manage to use it to reproduce the results from the WMT2020 shared task. We tried to reproduce their augmentation method (for which they did not release code), but obtained no improvement to performance.} 

Since this is the first work on PreQuEL, there are no immediate candidates for baselines. An exception is discussed in \S\ref{QE eval} and App.~\S\ref{QE appendix}.
We take the negated length of the sentences as a baseline. We assume the longer a sentence is, the harder it is to translate. We experimented also with the perplexity score of GPT-2 (for English as the source language), resulting with low correlation ($0.06$). We discuss more correlated features in \S\ref{standard features}. 
In places where we compare ourselves to a QE system, we use TransQuest. Also here we adjust the batch-size to 4 to fit our GPUs.

%%%%%%%%%%%%%%%%%%%%%%%%%%%%%%%%%%%%%%%%%%%%
\section{Main Results}\label{results}

In Table~\ref{table:main_res}, we present the main results. All of our models outperform the baseline. Similar to the case in QE, en-de is more challenging than et-en.
The augmentation improves the results in both settings -- intertraining (\textsc{Simple} Aug) and multi-task (\textsc{Multitask}), confirming the value of our augmentation method. \textsc{Multitask} outperforms the other architecture, with an improvement of more than 14 points over \textsc{Simple}, suggesting that different metrics capture different properties that can be useful for PreQuEL. As for \textsc{Combined}, \textsc{Combined}+ (with parser) outperforms \textsc{Combined}- (same size model, without parser) by more than 6 points, confirming the value of the syntactic knowledge, and not just the model size. \textsc{Combined}+ outperforms \textsc{Simple} Aug by 1 point. However, this improvement was achieved at the cost of doubling the size of the model, and is outperformed by the much smaller \textsc{Multitask}.

We use TransQuest as our upper-bound, to compare the performance of a similar QE model. \textsc{Multitask} is outperformed by TransQuest\footnote{Despite using the official code and our best efforts, TransQuest performance is lower than reported in its paper.} by 4.5 points.

%\footnote{We use the code from their github. We did not manage to use it to reproduce the results from the WMT2020 shared task. We tried to reproduce their augmentation method (for which they did not release code), but obtained no improvement to performance.} 

\begin{table}[t!] \footnotesize
\centering
    \begin{tabular}{ l||c|c} 
    \hline
    \multicolumn{1}{l}{} & \multicolumn{1}{c}{en-de} & \multicolumn{1}{c}{et-en} \\
    \hline
    \multicolumn{3}{l}{\textbf{PreQuEL Models}}\\
    \hline
    {\textsc{Simple}} &  0.196$\pm$0.02 & 0.602$\pm$0.00 \\ 
    {\textsc{Simple} Aug} & 0.315$\pm$0.00 & - \\ 
    {\textsc{Combined}+} & 0.326$\pm$0.02 & - \\ 
    {\textsc{Combined}-} & 0.265$\pm$0.00 & - \\ 
    {\textsc{Multitask}} & \textbf{0.336}$\pm$\textbf{0.01} & - \\ 
    {Baseline} & 0.135 & 0.050 \\ 
    \hline
    \multicolumn{3}{l}{\textbf{Upper Bound (QE)}}\\
    \hline
    {TransQuest} & 0.381$\pm$0.04 & 0.767$\pm$0.00 \\ 
    % {Transquest aug} & - & - \\ 
    % {Transquest pre-train} & 0.46916 & - \\ 
    % {Transquest aug pre-train} & - & - \\ 
    \hline
    \end{tabular}
\caption{Pearson’s $r$ of the predictions of the PreQuEL models with the DA scores, for en-de and et-en. All our models outperform the baseline, and the augmentation improves the results both as intertraining and multi-task. The et-en pair correlation is much higher than the correlation of en-de.}
\label{table:main_res}
\end{table}

\section{Analysis and Discussion}

% \subsection{DA vs. COMET}

% % \begin{table}[t!] \footnotesize
% % \centering
% %     \begin{tabular}{ c||c|c } 
% %     \hline
% %     {Model} & {DA} & {COMET} \\
% %     \hline
% %     {\textsc{Simple}} & $0.196\pm0.024$ & $0.219\pm0.005$ \\ 
% %     \hline
% %     \end{tabular}
% % \caption{Pearson’s correlation of the predictions of two \textsc{Simple}s: one trained on En-De DA data, and one on En-De COMET data created from the WMT NewsTests.}
% % \label{table:da-vs-comet}
% % \end{table}

% To examine the differences between the DA and COMET data, we want to conduct a fair comparison of them. We control the size of the training set by randomly choosing 7,000 sentences (the number of sentences in the DA train) from the en-de NewsTests training set. We train a \textsc{Simple} on this small NewsTests train, and compare its performances to the one of the \textsc{Simple} that was trained on en-de DA.
% We run each ensemble 3 time with different seeds, to confirm the results. The DA model gets an average correlation of $0.196$ with std of $0.024$. The COMET model gets an average correlation of $0.219$ and std $0.005$. \sdy{Running the COMET model on the full data results with correlation of  $0.652$.}
% These results suggest that although the COMET score seems to be easier to predict, the size of the training dataset is still the most important factor.
% % if we had COMET and DA scores for at least part of the test it was super cool.
% % add the results of the COMET tests? I think we should.

\subsection{General state-of-the-art MT Systems}
\label{Generalizing to Other MT Systems}
%\oa{we need a name for this variant, and then we can just write that and refer to it from sec 2}

%\oa{as I've written above, I would address this issue right from the start. define two variants of the task.}

So far, we showed results for the first variant of the task, namely for predicting the quality of outputs that were created using a single known MT system. 
%The translations of the gold labels were produced with publicly available data and state-of-the-art models trained the fairseq toolkit \citep{ott-etal-2019-fairseq}.
Here we approach the second variant, the one that predicts the score for a general state-of-the-art MT system. We expect a PreQuEL model for this variant to predict the quality score for any state-of-the-art MT system, without any supervision of it. Our experiments with this variant confirm our assumption. Some shared properties make a sentence easier or harder to translate, across systems.

%\oa{start differently: say that we do not have manual data for other systems as DA data is generally scarce. however, and given the correlation between DA and reference-based metrics, we can create approximate data using the pipeline...}
Manual DA annotations are scarce, and there is no data for other systems except for the WMT's DA of the MT task participants \citep{barrault-etal-2020-findings}, which is partial and small. Given the correlation between DA and reference-based metrics, we approximate the data using the augmentation pipeline described in section \ref{Automatic Evaluation Metrics Data}, this time translating with Facebook FAIR \citep{ng-etal-2019-facebook}.
The correlation of the train COMET labels of Marian with the train COMET labels of Facebook FAIR ranges from 0.60 to 0.82, depending on the dataset. This result implies that although there are differences in the translation quality, there is a lot in common too. To test what of this similarity our model catches, we take a \textsc{Simple} that was trained on data created with Marian and test it on data created with Facebook FAIR. 

The correlation on the Marian test set is $0.652$, while the correlation on the Facebook FAIR test set is $0.610$. One might expect the PreQuEL model to be bounded by the correlation between MT outputs ($0.82\cdot0.652=0.535$). However, these results do not fit this hypothesis ($0.610 > 0.535$) and indicate that the PreQuEL model generalizes well to other systems. Everything above the point of similarity in MT systems predictions is evidence of the preference of PreQuEL towards shared cues for MT system performance.

\subsection{Word-Ordering}
\label{word ordering}
To further investigate the role of syntax, we use two German word-ordering challenge sets \citep{Choshen2021TransitionBG}.
These two datasets \footnote{Challenge sets are available in: \url{https://github.com/borgr/nematus}} consist of pairs of sentences, each pair consists of two sentences, both holding the same meaning. They differ only in the order of the subject and object, the first is in the more common order, subject before object, and the second is object before subject.

On the first dataset, it is the case that resolves the ambiguity and determines which is the subject and who is the object. For example, \emph{``Das Kind bringt den Ball''} and \emph{``Den Ball bringt das Kind''} should be both translated to \emph{``The child brings the ball''}. 

On the second dataset, it is the verb that determines this. For example, \emph{``Die Katzen kicken die Maschine''} and \emph{``Die Maschine kicken die Katzen''} should be both translated to \emph{``The cats kick the machine''}. Additionally, on the second dataset each pair has its reverse-pair, in which the subject and object switch their roles. So we also have \emph{"Die Maschine kickt die Katzen"} and \emph{"Die Katzen kickt die Maschine"}, where both should be translated to \emph{"The machine kicks the cats"}.

% We hypothesize that the predictions for the subject-object sentences will be higher than the predictions for the less common order object-subject sentences.
% \sdy{We are also interested in the correlation between the predictions of the four versions we have for each sentence in the second dataset: 1. subject-object, 2. object-subject, 3. reversed-meaning-subject-object, 4. reversed-meaning-object-subject. 
% For example, if the correlation between sentences with the same meaning (1-2 and 3-4) will be higher than the correlation between sentences with the same syntax (1-3 and 2-4), we would conclude that meaning plays a more important role for our model than the syntax. However, if the correlation between sentences with the same syntax (1-3 and 2-4) will be higher that the correlation between sentences with the same structure (1-4 and 2-3), then we would conclude that the syntax is more important for our model than the structure.} 

% We train a PreQuEL model on German data. We select English to be the target language because it does not allow the flexibility in word-ordering that German does. The translations to English of both sentences in each pair should be the same. As in \S\ref{Learning from the Reference}, we create COMET German-English data by 'reversing' the English-German data. We train a \textsc{Simple} on it, and use the model to predict the scores of the datasets.
% We use the de-en \textsc{Simple} COMET model to predict the scores of the dataset.
Similar to \S\ref{Generalizing to Other MT Systems}, we approximate the data with the augmentation to allow data for training de-en \textsc{Simple}. We use it to predict the scores of the dataset.
We measure the correlation between the predictions of the four versions we have for each sentence in the second dataset: 1.subject-object, 2.object-subject, 3.reversed-meaning-subject-object, 4.reversed-meaning-object-subject. The correlation between sentences with the same meaning (1-2 and 3-4) is higher than the correlation between sentences with the same syntax (1-3 and 2-4) or sentences that are more similar in terms of their linear word order (1-4 and 2-3). For the full results see App.~\S\ref{word ordering appendix}. We conclude that meaning plays a more important role for our model than the syntax and the linear word order.

We compare the mean predictions of the subject-object order with the object-subject order. In the first dataset the mean score for the more common order subject-object is higher. In the second dataset however, for the reversed pair subject-object is indeed higher, but for the non-reversed pair the object-subject is the higher (See App.~\S\ref{word ordering appendix}). %3.reversed-meaning-subject-object is indeed higher than 4.reversed-meaning-object-subject, but 2.object-subject is higher than 1.subject-object.
% We conduct a paired t-test to determine if the sentences in the subject-object order receive higher predictions. We reject the null hypothesis for the first data set\oa{p-value?} \sdy{running it again.. waiting for the final results}, and only for the reversed pairs of the second dataset. The non-reversed pairs get on average higher predictions for the object-subject order.  

%\sdy{Evidently, the correlation between same-meaning sentences is higher than the correlation between same-syntax sentences. We conclude that our model relies more on the meaning, hinting semantics rather than syntax may also be what makes sentences hard to translate.}

\subsection{The Model as an Analytic Method}

Following the results of the word-ordering challenge set, we consider the possibility of using the model as an analytic method.
Concretely, we compare the PreQuEL scores for the source sentence, and a modified version of it (e.g., in past tense). 
This allows inspection of the effect linguistic aspects of the source have on translation quality.

We conduct our analysis on the following transformations proposed and implemented by NL-Augmenter repository \citep{dhole2021nlaugmenter}: {\it GenderSwap}, {\it TenseTransformation}, {\it RandomDeletion}, {\it YesNoQuestionPerturbation}, {\it ChangePersonNamedEntities}, {\it MultilingualBackTranslation}, {\it ReplaceNumericalValues}, {\it YodaPerturbation}.\footnote{The computationally demanding nature of these experiments prohibit us from using the entire set of transformations.} See App.~\S\ref{transformations appendix}.

Transformations might not affect some sentences. In the {\it TenseTransformation-past} for example, if the sentence is already in the past tense, it shouldn't change. Therefore, we report the number of sentences that were changed, and the mean score before and after the transformation for this subset of sentences.
For prediction, we use en-de \textsc{Simple} Aug. 

The transformations we use are automatic. Hence, some level of noise is expected, which may lower scores. Nevertheless, Table \ref{table:aug} shows that there are cases where the difference between the original sentences and the transformations is relatively small, and cases where it is more substantial. For example, {\it ChangePersonNamedEntities} and {\it ReplaceNumericalValues} show a difference smaller than 0.01, in agreement with our expectations. That is, we would not expect a change of a personal name or a numerical value to affect the performance of an MT system. On the Other hand, {\it RandomDeletion} and {\it YodaPerturbation} show a bigger difference, in agreement with the more drastic changes they make.

Examining {\it TenseTransformation}, we see that the past and present are similar, but the future is perceived to be harder to translate. This might be due to the future tense being less common.

Previous work suggested that translationese renders a text simpler \citep{cb567c06c5bd4728a25f5b5cead318be}, or at least different \citep{rabinovich-wintner-2015-unsupervised} than the source text. Table~\ref{table:aug} shows that when translating the sentences to German and then back to English, their mean score is indeed higher, rendering back-translation easier to translate.

\begin{table}[t] \footnotesize
\centering
    \begin{tabular}{ l||c|c|c|c } 
    \hline
    {Transformation} & {\# sent} & {src} & {trans} & {diff} \\
    \hline
    {GenderSwap} & 324 & 73.72 & 73.64  & -0.08 \\ 
    {RandomDeletion} & 975 & 73.77 & 73.15 & \textbf{-0.62} \\
    {YesNoQuestion} & 553 & 73.83 & 73.51 & \textbf{-0.32} \\
%    {PunctuationWithRules} & 78 & 73.66 & 73.68  & +0.02 \\
    {ChangePersonName} & 148 & 73.78 & 73.74 & -0.03 \\
    {BackTranslation} & 996 & 73.78 & 73.89 & \textbf{+0.11} \\
    {ReplaceNumericalVals} & 127 & 73.54 & 73.51 & -0.03 \\
    {YodaPerturbation} & 955 & 73.79 & 72.94 & \textbf{-0.85} \\
    {Tense-past} & 403 & 73.78 & 73.64 & \textbf{-0.16} \\
    {Tense-present} & 752 & 73.73 & 73.65 & -0.08 \\
    {Tense-future} & 904 & 73.76 & 73.50 & \textbf{-0.26} \\
    \hline
    \end{tabular}
\caption{Mean predictions of a \textsc{Simple} Aug before and after applying transformations. Differences greater than 0.1 are. boldfaced.}
\label{table:aug}
\end{table}

\subsection{Different Target Languages}
\label{Different Target Languages}
\begin{table}[t!] \footnotesize
\centering
    \begin{tabular}{ c||c|c } 
    \hline
    {Model} & {en-de DA} & {en-zh DA} \\
    \hline
    {\textsc{Simple} en-de} &  \textbf{0.377} & 0.260 \\ 
    {\textsc{Simple} en-zh} & 0.140 & \textbf{0.577} \\ 
    \hline
    \end{tabular}
\caption{Pearson’s $r$ of the predictions of two \textsc{Simple}s Aug: one trained on en-de data and one on en-zh, with the matching DA scores for both en-de and en-zh. Both models have a higher correlation with the gold labels of the development set of the language that they were trained on. }
\label{table:3}
\end{table}

\begin{table*}[t] \footnotesize
\centering
    \begin{tabular}{ c||c|c|c|c|c|c|c|c|c|c } 
    \hline
    { } & {length} & {depth} & {LM} & {VERB} & {advcl} & {case} & {unigram} & {bigram} & {3gram} & {4gram} \\
    \hline
    {model preds} & \textbf{-0.2894} & \textbf{-0.2157} & \textbf{0.2151} & \textbf{-0.3144} & \textbf{-0.2541} & \textbf{-0.2056} & \textbf{0.2230} & \textbf{0.3258} & \textbf{0.3362} & \textbf{0.3338} \\ 
    {gold labels} & -0.1305 & -0.0338 & 0.0968 & -0.1424 & -0.0848 & -0.0822 & 0.1045 & 0.1546 & 0.1638 & 0.1625 \\ 
    \hline
    \end{tabular}
    % {\color{red} first say what this table presents and what the rows/cols mean}
\caption{Pearson’s $r$ of \textsc{Simple} Aug predictions and the gold labels of the en-de DA dev, with the the selected features. The correlation with the features is higher for the model predictions than for the gold labels, implying that our model overestimates the importance of the features. }
\label{table:4}
\end{table*}

The syntactic and semantic structures converge and diverge between different languages, and there are therefore cases in which the translation of one language into another results in a very different structure than that of the source \citep{nikolaev-etal-2020-fine, dorr-1994-machine}. Does PreQuEL capture hard sentences in the seen source, or does it implicitly capture divergences between the source and target languages?

To assess this we consider en-de and en-zh pairs. Like en-de, en-zh is both high resourced and obtains high QE results \citep{specia-EtAl:2020:WMT2}. The correlation of their DA scores is 0.08, meaning that despite having the same input sentences, what is considered hard is quite different.
%This might be due to the amount of resources that were put in developing the MT systems for these languages, or due to the closeness of one of the target languages to the source language.

In the PreQuEL settings on the other hand, the model is exposed only to the source sentences. The target sentences are only indirectly exposed by the DA score. Would it be enough for the model to learn what is hard to translate to one language, but not to another?

We use instances of \textsc{Simple} Aug for en-de and en-zh to predict the scores for both en-de and en-zh development DA.
We compare the correlation of the models' predictions with each other and the correlation of each model with the gold scores of each target language.
% add table

In Table \ref{table:3}, we see that both models have a higher correlation with the gold labels of the development set of the language that they were trained on. We conclude that the models manage to learn a function that is specific not only to the source language but also to the target language. The model predicts the difficulty of translating a sentence from a specific language to a specific language, and not just the general difficulty of the sentence.
However, the correlation between the predictions of the models with each other is $0.588$, much higher than the correlation between the DA labels. Thus, compared to the DA labels, the PreQuEL model does tend to overestimate the source.

\subsection{Correlating with Standard Features}
\label{standard features}

We further examine the correlation between our model predictions and standard NLP features. We consider sentence length, Universal Dependencies (UD) \citep{nivre-etal-2020-universal} parse tree depth, number of edges with a specific UD label, number of tokens with a specific POS label, language detector confidence, language model score, and n-gram model probabilities. See App.~\ref{extract-nlp-features-appendix}.

We report only the features that are statistically significant ($P<0.0006$) after Bonferroni correction and that show at least 0.2 correlation with the predictions or the labels -- sentence length, tree depth, language model score, verb POS count, adverbial clause label count, case label count, and n-grams.

In Table~\ref{table:4}, we can see that the correlation with the features is higher for the model predictions than for the gold labels. This implies that our model not only uses these features, but also over-estimates their importance.

%%%%%%%%%%%%%%%%%%%%%%%%%%%%%%%%%%%%%%%%%%%%%
% Omri here 14/3
\section{Determining whether to MT}

We expect that the standard test case for PreQuEL would be to determine the chances of a sentence to be translated correctly to its target language. According to the FLORES guidelines, a translation that closely preserves the semantics of the source sentence gets a DA score of 70-90. Only a perfect translation gets a score of 91-100. Therefore, we report the Precision/Recall curve for a threshold of 70, to find the 'good enough' sentences, following \cite{guzman-etal-2019-flores}. In App.~\ref{determining-whether-to-mt-appendix} we also report a threshold of 90 to find the perfect ones.

\begin{figure}[t]
\centering
\includegraphics[width=8.6cm]{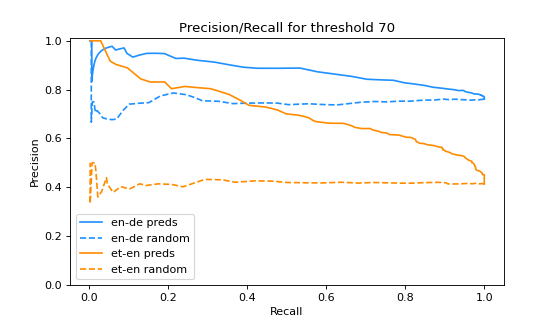}
\caption{Precision/Recall curve for en-de and et-en development datasets, for the prediction of whether a DA threshold of over 70 would be given to the translation of the input sentence. The blue plots are comparable and so are the orange (but not cross-color comparisons).}
\label{figure:pr-threshold-70}
\end{figure}

We use the predictions of \textsc{Simple}s Aug to estimate what sentences will get a DA score above 70. Figure \ref{figure:pr-threshold-70} shows the Precision/Recall curves for en-de and et-en, for a DA threshold of 70. For comparison, we plot the curves for a random model that uniformly predicts a random score between 0-100 for each sentence. For en-de, 76\% of the sentences have a DA score above 70, so we can maintain a relatively high precision simply by assigning random scores. Still, our model outperforms it. For et-en, 43\% of the sentences are above 70.

% \begin{figure}[t]
% \centering
% \includegraphics[width=7.5cm]{images/PR_threshold_70_et-en.png}
% \caption{Precision/Recall curve for et-en dev data, with DA threshold 70}
% \label{figure:3}
% \end{figure}

%%%%%%%%%%%%%%%%%%%%%%%%%%%%%%%%%%%%%%%%%%%%%
\section{Quality Estimation}
\label{QE}

\subsection{Augmentations for Quality Estimation}
\label{QE aug}

As mentioned before, our data augmentation method might be useful for training QE systems too. QE still has no access to the reference that metrics rely on. Extracting similarities between the translation to a hidden reference might force new abilities to emerge in the QE networks.

%To test that, and to provide an 'upper bound' to our performance, 
We train TransQuest on en-de DA data, with and without intertraining on COMET data. We run each version 3 times with different seeds, to confirm the results.
%We use the released implementation, changing only the batch size from 8 to 4 due to limited resources. We ensemble with 3 runs. 
The average correlation for TransQuest without the augmentations is $0.381$ with std $0.043$  
%\footnote{In the WMT2020 shared task TransQuest performed better. They used their own augmentation method, which they describe in the paper but don not release its code. We tried to reproduce it, apparently with no success because it hurt the performance.}. 
The average correlation for TransQuest with the augmentations is $0.429$ with std $0.008$. TransQuest thus benefits from the augmented data, in terms of both correlation and stability (much smaller std).

\subsection{Quality Estimation Evaluation}
\label{QE eval}
\citet{sun-etal-2020-estimating} argue that the recent success of pre-trained language models for QE is over-stated. 
%Although to the best of our knowledge our work is the first to address PreQuEL, Sun et al. did 
They point out that it is possible to perform QE to a large extent using only the source or output sentence (we outperform their results; see  App.~\S\ref{QE appendix}).
%For the en-de pair for example, the correlation between predictions from a BERT based QE model with source sentences only and gold HTER labels was 94.6\% of its performance with full inputs.
%\lc{before this, start with what they look for. They did not care about prequel. their motivation was evaluation or looking at a dataset or something} 
They viewed this finding as an artifact of the dataset, as they expect the predictions of a well QE system to reflect both the translation's closeness to the source text, and how well it fits in the target language. The good results the model obtained with only source/translation imply that only one of the two is taken into account. They suggest to replace HTER with a metric that represents both fluency and adequacy, such as DA.
%, and ensure there are enough representation instances with high and low adequacy and fluency.

%We however argue that the pre-QE model can potentially simulate the MT system, and therefore there is no theoretical reason to consider the dataset as ``cheatable''.

% However, \citet{sun-etal-2020-estimating} showed that QE systems can perform well even when given only source or translated sentences. For the en-de pair for example, the Pearson's correlation between predictions from a BERT based QE model and gold HTER labels was 94.6\% of its performance with full inputs.
% Sun et al. suggest that this shows that these datasets are `cheatable', and that QE systems trained on them would not generalize well. They suggest to use a metric that represents both fluency and adequacy as labels, such as DA and ensure there are enough representation instances with high and low adequacy and fluency.

%While this result indeed implies that existing evaluation method for QE systems is problematic, because of the nature of our task we obviously do not treat these results them-self as a problem.
We interpret these results differently. Indeed, these experiments provide motivation for PreQuEL, showing that the source sentence holds considerable information required for predicting system performance. However, we argue that an oracle PreQuEL model can simulate the MT system, and therefore there is no theoretical reason to consider the dataset as ``cheatable''.

As for their suggestion to use DA, we find that their criticism is not specific to HTER. We train a \textsc{Simple} on en-de HTER data and compare its performance to the performance of a \textsc{Simple} that was trained on the en-de DA data.
We run each ensemble 3 times with different seeds, to confirm the results. The DA model gets an average correlation of $0.196$ with std of $0.024$. The HTER model gets an average correlation of $0.322$ and std $0.013$.
These results suggest that predicting the DA score from the source alone is indeed somewhat more difficult than predicting the HTER, but definitely possible.
%\sdy{Regrading Sun et al. original results, we outperforms their results by 0.02 points.}

However, we agree that QE systems should not ignore parts of their input, and that this should be addressed in evaluation. QE systems in the WMT shared task are trained and evaluated against the outputs of a single MT system. Possibly, the QE systems learn to simulate it, which would explain the redundancy of the translation.

Instead, we suggest to train and test the QE models on datasets of multiple MT systems. Such a dataset would 
%enforce the use of both source and translation sentences. This can be done with datasets that 
include translations from diverse systems. This would sever the ability of the QE systems to simulate the translation process implicitly. Therefore, QE systems would have to rely on the actual translation for successful prediction.

Moreover, to assert reliance on the source, translation of unrelated sentences should be included in the translation as well.
% what about making the source crucial?

% \begin{table}[t!] \footnotesize
% \centering
%     \begin{tabular}{ c||c|c } 
%     \hline
%     {Model} & {DA} & {HTER} \\
%     \hline
%     {\textsc{Simple}} &  $0.196\pm0.024$ & $0.322\pm0.013$ \\ 
%     % & cell8 & cell9 \\ 
%     \hline
%     \end{tabular}
% \caption{Pearson’s correlation of the predictions of two \textsc{Simple}s: one trained on En-De DA data and one on En-De HTER data, with the matching DA/HTER labels.}
% \label{table:da-vs-hter}
% \end{table}

\section{Conclusion}

We presented a new task, PreQuEL, the task of predicting the quality of the output of MT systems based on the source sentence only. We developed a baseline model and reported its results, providing motivation for the task by showing that considerable information for predicting the quality scores is stored in the source sentence. We developed an automatic augmentation method, and used it to improve our results. We showed that the predictions made by a model that was trained with a specific system supervision do generalize to other state-of-the-art systems. We analyzed our model by testing it on challenge sets and other languages, concluding that our model is aware of syntax, meaning, and the target language.

Motivated by these latter results, we suggested to use the PreQuEL model as an analytic method, to confirm the effect of linguistic and semantic phenomena on the ability of a MT system to translate. Future work will use this method to provide insights into the performance boundaries of current MT systems. Other lines of work we intend to pursue include examining the advantages of a multilingual PreQuEL model, and developing an advanced PreQuEL model that selects the MT system that is most likely to generate the best translation for a given text.

\section{Limitations}
Although our PreQuEL model for the en-de language pair reaches good results when compared to the TransQuest upper bound, the correlation is still not high in absolute terms ($0.336$ \S\Ref{results}). %\oa{what is the IAA? add this as a ref point} \sd{They didn't report it...}
Therefore, its predictions should be used with care.

One of our motivations in this paper is to investigate whether there are any linguistic features, that make a sentence more difficult for MT. We trained models for the task and reported their performance, supporting the claim that there are such features. However, trained with an end-to-end approach, our models are not suitable for explicitly pointing out these features. That is, when our models predict a sentence to be hard to translate, we can not tell why.
We did however show some properties that our models are sensitive to. For example, the \textsc{Combined} architecture (\S\Ref{Advanced Models}) confirmed the role of syntax, and the German word ordering challenge sets (\S\Ref{word ordering}) revealed the importance of meaning. We also showed a correlation with standard features, to demonstrate their influence.%\oa{this last sent is not clear} 

Another limitation is the available data. In some of the analysis experiments we needed data for languages pairs that do not appear in the shared task on QE \citep{specia-etal-2021-findings}, or translation outputs coming from multiple MT systems. In these cases, we used automatically augmented data (\S\Ref{Automatic Evaluation Metrics Data}), instead of proper DA labels. This data is naturally of lower quality. We note that this limitation refers to some of the analysis only, and not our main results, which we tested on DA data only.

\section*{Acknowledgments}
We thank Anna Pellivert and Menachem Shefer for helpful discussions. This work was supported in part by the Israel Science Foundation (grant no. 2424/21), and by the Applied Research in Academia Program of the Israel Innovation Authority.

\bibliography{main}
\bibliographystyle{acl_natbib}

\appendix

\section{Computing Infrastructure}
\label{Computing appendix}

Our architecture is roberta-large based, and therefore the number of parameters for each instance of our model is ~355M$\times$3 due to ensembling. The \textsc{Combined} architecture uses two instances of roberta-large, and therefore is ~355M$\times$3$\times$2. 

When training a PreQuEL model with a source language that is not English, we replace roberta-large with xlm-roberta-large \citep{DBLP:journals/corr/abs-1911-02116}. 

We train each instance of our PreQuEL models on 2 CPU and 1 GPU. The run-time was 2.5 hours for a training set of 7K sentences. The \textsc{Combined} architecture with intertraining took the longer to train, ~55 hours.

\section{Extracting Standard NLP Features}
\label{extract-nlp-features-appendix}

To extract the language detector confidence we use spacy. To extract the n-grams we use kenlm (\url{https://github.com/kpu/kenlm}). To extract POS tags, UD labels and UD tree depth we use stanza \citep{qi2020stanza}. To extract language model score we use lm-scorer (\url{https://github.com/simonepri/lm-scorer}).  

\section{Additional Experiments}
\subsection{DA vs. COMET}
\label{DA vs. COMET}

% \begin{table}[t!] \footnotesize
% \centering
%     \begin{tabular}{ c||c|c } 
%     \hline
%     {Model} & {DA} & {COMET} \\
%     \hline
%     {\textsc{Simple}} & $0.196\pm0.024$ & $0.219\pm0.005$ \\ 
%     \hline
%     \end{tabular}
% \caption{Pearson’s correlation of the predictions of two \textsc{Simple}s: one trained on En-De DA data, and one on En-De COMET data created from the WMT NewsTests.}
% \label{table:da-vs-comet}
% \end{table}

To examine the differences between the DA and COMET data, we want to conduct a fair comparison of them. We control the size of the training set by randomly choosing 7K sentences (the number of sentences in the DA train) from the en-de NewsTests training set. We train a \textsc{Simple} on this small NewsTests train, and compare its performances to the one of the \textsc{Simple} that was trained on en-de DA.
We run each ensemble 3 time with different seeds, to confirm the results. The DA model gets an average correlation of $0.196$ with std of $0.024$. The COMET model gets an average correlation of $0.219$ and std $0.005$. Running the COMET model on the full data results with correlation of  $0.652$.
These results suggest that although the COMET score seems to be easier to predict, the size of the training dataset is still the most important factor.
% if we had COMET and DA scores for at least part of the test it was super cool.
% add the results of the COMET tests? I think we should.

\subsection{Out-Of-Domain Datasets}
\label{out-of-domain}

As discussed in \S\ref{exp setup}, our datasets differ on their domain. We take instances of \textsc{Simple}, with no ensembles this time (to reduce training time), and train them on ChrF++ augmentations. First, to measure in-domain, we train and test one instance on each one of the dataset. To measure out-of-domain, for each dataset we train another instances, this time avoiding this dataset during training, and only testing on it (e.g. train on NewsTests, bible-uedin and GlobalVoices, test on Tatoeba). We use the development sets for the testing.

We present the results in Table~\ref{table:ood}. The correlations vary between the datasets, both for the in-domain and for the out-of-domain. We conclude that similar to other NLP tasks, the domain plays an important role.     

\begin{table}[t!] \footnotesize
\centering
    \begin{tabular}{ c||c|c|c } 
    \hline
    {} & {\#sents} & {in domain} & {out of domain} \\
    \hline
    {NewsTests} &  {28,887} & 0.64 & 0.30 \\
    {bible-uedin} & {48,705} & 0.72 & 0.36 \\
    {GlobalVoices} & {55,822} & 0.55 & 0.35 \\
    {Tatoeba} & {197,381} & 0.36 & 0.25 \\
    \hline
    % \hline
    \end{tabular}
\caption{The correlations vary between the datasets, both for the in-domain and out-of-domain. Pearson's $r$ of the predictions of the models with/without training on the dataset.}
\label{table:ood}
\end{table}

\subsection{Learning from the Reference}
\label{Learning from the Reference}

All of our experiments were focused on the ability of our PreQuEL model to give predictions on the expected quality of the translated sentence given the source sentence only. Can it do the same given the reference sentence instead?

In cases where we have a good quality reference (e.g., parallel corpus) we assume the source and reference sentences hold the same content, so it makes sense to expect the answer to this question to be yes. However, it is possible that the challenge of translating from one language to another is more directly related to some features on the source side. 

To test that without the artifact of the input language, we use de-en COMET data (See \S\ref{word ordering}). This way we will have our references in English.

The results for one de-en \textsc{Simple} COMET instance that was trained on references is $0.639$, similar to a model that was trained on the source.

\subsection{Outperforming Existing 'PreQuEL' Model}
\label{QE appendix}
We train and test \textsc{Simple} Aug on the dataset that Sun et al. used, the QE dataset from WMT2019. This dataset uses HTER to score the quality. They reported a correlation of $0.400$, while we mange with intertraining on COMET to achieve correlation of $0.422$. 

The QE dataset from WMT2019 containes $13,442$ training samples, much more than the 7k of the WMT2020. Therefore, although we did improve the results over theirs, the augmentation gain is smaller than we showed in \S\ref{results} for the WMT2020 DA.   

\section{Word Ordering full results}
\label{word ordering appendix}

Tables~\ref{table:word-ordering-mean} presents the mean scores for both datasets. Table~\ref{table:word-ordering-mean-2} present the correlation between all sentence versions of the second dataset.

\begin{table}[t!] \footnotesize
\centering
    \begin{tabular}{ l||c} 
    % \hline
    % \multicolumn{1}{l}{} & \multicolumn{1}{c}{en-de} & \multicolumn{1}{c}{et-en} \\
    \hline
    \multicolumn{2}{l}{\textbf{First Dataset}}\\
    \hline
    {1.sub-obj} & $0.845$ \\ 
    {2.obj-subject} & $0.799$ \\ 
    \hline
    \multicolumn{2}{l}{\textbf{Second Dataset}}\\
    \hline
    {1.sub-obj} &  $0.865$ \\ 
    {2.obj-sub} & $0.870$ \\ 
    {3.re-sub-obj} & $0.869$ \\ 
    {4.re-obj-sub} & $0.851$  \\ 
    % \hline
    % \multicolumn{2}{l}{\textbf{Second Dataset reversed}}\\
    % \hline
    % {3.re-sub-obj} & $0.86$ \\ 
    % {4.re-obj-sub} & $0.851$  \\ 
    \hline
    \end{tabular}
\caption{Mean score predictions for the syntax datasets.}
\label{table:word-ordering-mean}
\end{table}

\begin{table}[t!] \footnotesize
\centering
    \begin{tabular}{ l||c } 
    \hline
    {sents pair} & corr \\
    \hline
    {1.sub-obj with 2.obj-sub} & \textbf{0.901} \\
    {1.sub-obj with 3.rev-sub-obj} & $0.687$ \\
    {1.sub-obj with 4.rev-obj-sub} & $0.680$ \\
    {2.obj-sub with 3.rev-sub-obj} & $0.563$ \\
    {2.obj-sub with 4.rev-obj-sub} & $0.641$ \\
    {3.rev-sub-obj with 4.rev-obj-sub} & \textbf{0.936} \\
    \hline
    \end{tabular}
\caption{Correlation between all sentences versions, for the second dataset.}
\label{table:word-ordering-mean-2}
\end{table}

\section{NL-Augmenter transformations}
\label{transformations appendix}

% \url{https://github.com/GEM-benchmark/NL-Augmenter} 

\paragraph{GenderSwap}
This transformation swaps all gendered words in a given sentence with their counterparts. Names are also randomly swapped. For example "Bob wants to become a programmer, as his father" is transformed to "Alice wants to become a programmer, as her mother".

\paragraph{TenseTranformation}
This transformation converts sentences from one tense to the other, for example, "My father goes to gym every day" is transformed to "My father went to gym every day".

\paragraph{RandomDeletion}
This transformation randomly remove each word of a sentence or paragraph with a probability p.

\paragraph{YesNoQuestionPerturbation}
This perturbation turns English statements into yes-or-no questions. For example, "He also begins an affair with Veronica Harrington, who bails him out." is transformed to "Does he also begin an affair with Veronica Harrington, who bails him out? Yes."

\paragraph{ChangePersonNamedEntities}
This transformation acts like a perturbation which changes the name of the person. For example, from "John" to "Cathy".

\paragraph{MultilingualBackTranslation}
This transformation translates a given sentence from a given language into a pivot language and then back to the original language.

\paragraph{ReplaceNumericalValues}
This transformation looks for numerical values in the text and replaces it with another random value of the same cardinality. For example, "6.9" may be replaced by "4.2", or "333" by "789".

\paragraph{YodaTransformation.}
This transformation modifies sentences to flip the clauses such that it like "Yoda Speak". For example, "You still have much to learn" is transformed to "Much to learn, you still have".

\section{Determining Whether to MT - Threshold 90}
\label{determining-whether-to-mt-appendix}

\begin{figure}[t]
\centering
\includegraphics[width=8.6cm]{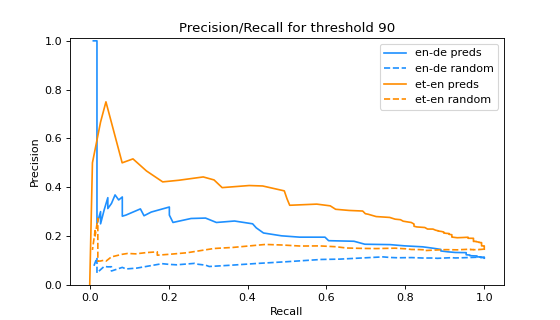}
\caption{Precision/Recall curve for en-de and et-en development datasets, for the prediction of whether a DA threshold of over 90 would be given to the translation of the input sentence. The blue plots are comparable and so are the orange (but not cross-color comparisons).}
\label{figure:pr-threshold-90}
\end{figure}

Figure \ref{figure:pr-threshold-90} presents the Precision/Recall curve for both en-de and et-en development datasets, with threshold 90.

\end{document}